\documentclass[letterpaper, 10 pt, conference]{ieeeconf}
\IEEEoverridecommandlockouts 
\usepackage[letterpaper, left=54pt, right=54pt, bottom=54pt, top=54pt]{geometry}

\usepackage{graphicx}
\usepackage{amsmath} 
\usepackage{amssymb}  
\usepackage{mathtools}
\usepackage{bbm}
\usepackage{dsfont}
\usepackage{multirow}
\usepackage[dvipsnames]{xcolor}
\usepackage{subfig}
\usepackage[noadjust]{cite}
\usepackage{enumerate}
\usepackage{booktabs}
\usepackage{hyperref}

\usepackage{algpseudocode}
\usepackage{algorithm}
\usepackage{caption}
\usepackage[free-standing-units=true]{siunitx}
\usepackage{cleveref}

\title{\LARGE \bf
Hierarchical Intention Tracking for \\ Robust Human-Robot Collaboration in Industrial Assembly Tasks}

\author{Zhe Huang*, Ye-Ji Mun*, Xiang Li$\dagger$, Yiqing Xie$\dagger$, Ninghan Zhong$\dagger$, \\ Weihang Liang, Junyi Geng, Tan Chen, and Katherine Driggs-Campbell
\thanks{* denotes equal contribution as the first author. $\dagger$ denotes equal contribution as the second author.}%
\thanks{Z. Huang, Y. Mun, X. Li, Y. Xie, W. Liang, and K. Driggs-Campbell are with the Department of  Electrical and Computer Engineering at the University of Illinois at Urbana-Champaign. emails: \{zheh4, yejimun2, xiangl5, yiqingx2, weihang2, krdc\}@illinois.edu}%
\thanks{N. Zhong is with the Department of Computer Science at the University of Illinois at Urbana-Champaign. email: ninghan2@illinois.edu}%
\thanks{J. Geng is with the Department of Aerospace Engineering at Pennsylvania State University. email: jgeng@psu.edu}%
\thanks{T. Chen is with the Department of Electrical and Computer Engineering at Michigan Technological University. email: tanchen@mtu.edu}%
\thanks{This work was supported by Foxconn Interconnect Technology through the UIUC Center for Networked Intelligent Components and Environments (C-NICE).}%
}

\begin{document}

\maketitle
\thispagestyle{empty}
\pagestyle{empty}

\begin{abstract}
Collaborative robots require effective human intention estimation to safely and smoothly work with humans in less structured tasks such as industrial assembly, where human intention continuously changes. We propose the concept of intention tracking and introduce a collaborative robot system that concurrently tracks intentions at hierarchical levels. The high-level intention is tracked to estimate human's interaction pattern and enable robot to (1) avoid collision with human to minimize interruption and (2) assist human to correct failure. The low-level intention estimate provides robot with task-related information. We implement the system on a UR5e robot and demonstrate robust, seamless and ergonomic human-robot collaboration in an ablative pilot study of an assembly use case. Our robot demonstrations and videos are available at \href{https://sites.google.com/view/hierarchicalintentiontracking}{https://sites.google.com/view/hierarchicalintentiontracking}.
\end{abstract}
\section{Introduction}\label{sec:Introduction}

Collaborative robot solutions are being actively developed in industrial tasks including part sorting, tool delivery, precise positioning, and cooperative transportation~\cite{perez2015fast, cheng2020towards, wojtara2009human, alevizos2020physical}. Teaming up humans and robots boosts production efficiency by combining cognition and dexterity from humans with repeatability and load carrying capacity from robots~\cite{wang2017trends}. One key challenge for these solutions is human uncertainty~\cite{huang2020coarse}. We argue that reliable human intention estimation is a critical component of safe and seamless human-robot collaboration. 

\begin{figure}[t]
    \centering
    \includegraphics[width=0.95\linewidth]{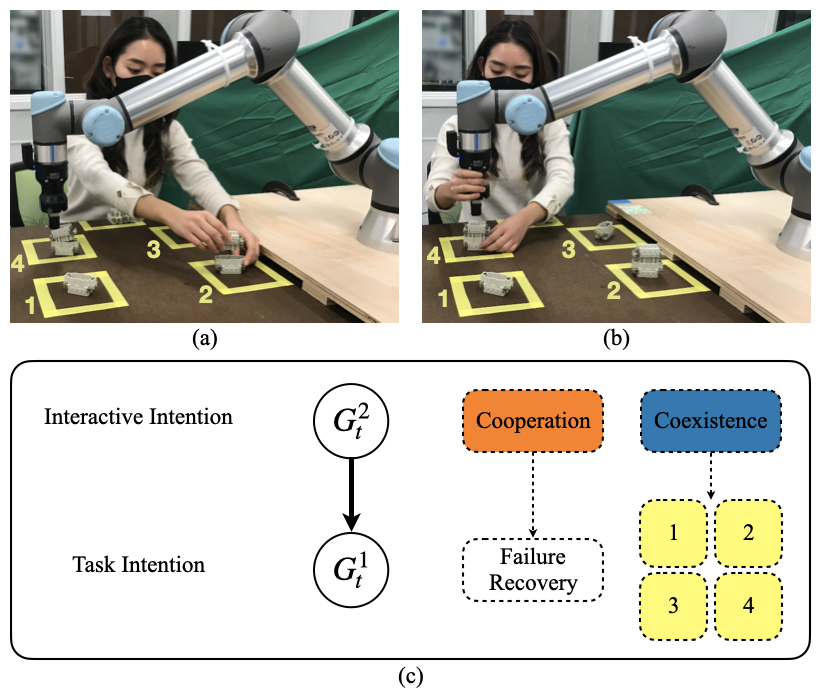}
    \caption{A person works with a robot to assemble parts in her preferred order. The robot must keep track of which part is her likely goal $G_t^1$, while estimating her interactive intention $G_t^2$. (a)~Coexistence mode: the robot performs pushing action to the parts she aligned while she is aligning other parts. (b)~Cooperation mode: the robot is manually guided to recover the failed pushing attempt. (c)~The hierarchy of human intention during collaboration.
    }
    \label{fig:intro}
    \vspace{-10pt}
\end{figure}

Many works have incorporated human intention estimation in highly structured settings~\cite{perez2015fast, villani2018survey}. 
We observe critical limitations when previous methods are applied to less structured tasks. Take a prototypical collaborative assembly task in Figure~\ref{fig:intro} as an example, where a human-robot team assembles four pairs of male and female parts at designated regions\footnote{This task is a less structured analogy to BMW/MINI Crash Can Assembly Task demonstrated at \url{https://youtu.be/keh99z1M5LI}, where human aligns rivets and robot performs rivet installation.}. The human is responsible for part alignment which requires delicate manipulation skills, and the robot is responsible for pushing male parts into female parts which requires a large force. In this application, the prior methods will fail due to two key factors: dynamic intention and intention hierarchy.

First, human intention changes at different stages of the task. The robot needs to estimate and follow the sequence by which the human aligns the parts. Prior works typically define human intention as a single random variable that does not evolve over time~\cite{cacace2018shared}. This definition transforms a multi-step task into multiple single-step tasks, and intention estimation can converge prematurely in one task before the next begins. Alternatively, the intention is inferred by single shot classification~\cite{driggs2015identifying}, but inference capabilities are limited with partial historical information.

Second, human intention is often composed of a multi-layer hierarchy. Previous works are focused on single-layer cases \cite{geravand2013human, nicolis2018human}, which can be inadequate to address unstructured settings. In the proposed example, a two-layer intention hierarchy is needed to entail efficiency and robustness. The high level includes \textit{coexistence interactive intention}, meaning the human intends to work without interruption from the robot, and \textit{cooperation interactive intention}, meaning the human intends to physically guide the robot. The low level includes which part the human intends to work on, and assembly failure correction.

In this work, we propose \textit{hierarchical intention tracking} to take these factors into account. We introduce a Hierarchical Intention Tracking (HIT) based human-robot collaboration system to simultaneously track both high-level interactive intention and low-level task intention. When the high-level estimate is coexistence, the robot performs collision avoidance while reaching for the task goal estimated at the low level. When the high-level estimate is cooperation, the robot approaches to the human and provides admittance control for failure recovery. We present two main contributions:

\begin{enumerate}[(1)]
    \item We derive \textit{intention tracking} based on a generic graphical model of intention-evolving human-robot collaboration by treating intention as a Markov process, and extend intention tracking to a multi-layer intention hierarchy.
    \item We develop a seamless, robust, and ergonomic human-robot collaboration system based on hierarchical intention tracking, which is demonstrated in a multi-step assembly task through ablative pilot study.
\end{enumerate}
\section{Related Work}\label{sec:related}

\subsection{Automation Modes in Human-Robot Collaboration}\label{sec:related-collaboration}
A variety of industrial applications involve human-robot collaboration such as tool handover~\cite{perez2015fast, cheng2020towards}, heavy object lifting~\cite{grahn2016potential, kim2017anticipatory}, surface polishing~\cite{wilbert2012robot}, welding~\cite{shi2012levels}, and assembly~\cite{tsarouchi2017human, heydaryan2018safety}. Human-robot collaboration has three major automation modes: safety, coexistence, and cooperation~\cite{de2012integrated, villani2018survey}. 
The safety mode enforces human and robot to not move at the same time in shared work space. Robot motion is paused when human is detected and is resumed after human leaves~\cite{svarny2019safe, papanastasiou2019towards}. The coexistence mode allows human and robot to work simultaneously in close proximity with no physical contact. Human and robot execute their own tasks, and prefer no interruption from the partner, so robot performs human avoidance~\cite{flacco2012depth,lasota2014toward, chen2018collision}. The cooperation mode offers physical coordination. Manual guidance and shared control are used to address situations which are beyond robot capabilities and require human intervention~\cite{geravand2013human,cacace2018shared, nicolis2018human}. To achieve robust and efficient close-proximity human-robot collaboration in complicated free-form assembly tasks, our work implements a coexistence module for concurrent operation and a cooperation module for failure recovery.

\subsection{Human Intention Estimation}\label{sec:related-intention}
The research on human-centered autonomy has extensively investigated the concept of human intention~\cite{li2013human,rasouli2019pie,katyal2020intent}. To estimate human intention, various types of observation on human behavior are used as input, including human trajectories~\cite{perez2015fast, rasouli2019pie, katyal2020intent}, gesture~\cite{tsarouchi2017human, mazhar2019real}, gaze~\cite{huang2016anticipatory}, speech~\cite{ahn2018interactive}, facial expressions~\cite{truong2019social}, and force-torque measurements~\cite{li2013human, cacace2018shared}. Many works study how human intention influences human behavior and benchmark approaches on human datasets~\cite{perez2015fast, rasouli2019pie}.
Other works consider the mutual influence between human and robot and take both human and robot states as input~\cite{wang2013probabilistic, park2016hi, cacace2018shared}. 

Human intention is typically represented by one fixed random variable. Thus, intention estimation is formulated either as intention recognition given a fixed length of observations~\cite{cacace2018shared, rasouli2019pie, katyal2020intent}, or recursive intention estimation, which produces an online maximum a posteriori over the same intention variable given tracked sequence~\cite{wang2013probabilistic, park2016hi, zanchettin2017probabilistic, du2020online}. Our previous work proposes intention mutation mechanism~\cite{huang2021long}, but the mechanism is developed in the context of pedestrian trajectory prediction, where only human behavior is considered and changing intention is regarded as anomaly. To resolve the premature convergence issue in recursive intention estimation approaches, in this work we generalize intention mutation mechanism to intention transition dynamics by treating intention as a state variable, and formally derive intention tracking in the context of human-robot collaboration.

Intention hierarchy is often discussed when human intention is defined in terms of commands. A command can be interpreted as a pyramid of a goal, sub goals, and primitives~\cite{hong2007mixed}. Hierarchical Task Network is used to exploit the hierarchical structure by generating candidates of flattened primitive sequences, and intention estimation still works at a single level to distinguish among candidates~\cite{cacace2018interactive}. Hierarchical Hidden Markov Model in~\cite{zhu2008human} recognizes intention sequences by incorporating two levels of Hidden Markov Models, but its higher level is used to refine results from the lower level through context awareness, where the intention is only defined at one level. Our work defines intention hierarchy in terms of abstraction levels which are not necessarily constrained to command or task related goals. Though implementation is based on the two-layer case in Figure~\ref{fig:intro}(c), our hierarchical intention tracking framework can be applied to an intention hierarchy with an arbitrary number of levels. In contrast to~\cite{hong2007mixed}, we apply Dynamic Bayesian Networks to study hierarchical intention tracking~\cite{murphy2002dynamic}, and Bayesian inference is performed both vertically (hierarchically) and horizontally (temporally).
\section{Hierarchical Intention Tracking}\label{sec:problem}

\begin{figure}[bt!]
    \centering
    \includegraphics[width=0.8\linewidth]{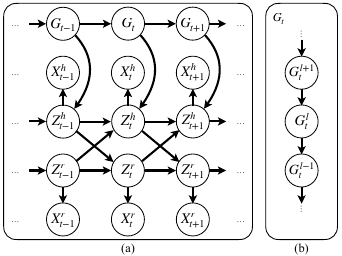}
    \caption{(a) A graphical model of intention-evolving human-robot collaboration. We denote observed human and robot states as $X^h_t$ and $X^r_t$. Latent human and robot states are represented by $Z^h_t$ and $Z^r_t$. The human intention is represented by~$G_t$. (b) A graphical model of hierarchical intentions, which is essentially a chain of intentions. The lowest-level intention $G_t^{1}$ is a parent of $Z^h_t$ in (a).}
    \label{fig:graphical-hierarchical}
    \vspace{-10pt}
\end{figure}

A human-robot team is assigned $m$ task goals. The human leads the team to accomplish all goals in his/her desired task sequence unknown to the robot. Human-robot team dynamics are formulated as the graphical model presented in Figure~\ref{fig:graphical-hierarchical}(a). The human intention $G_t$ is a Markov process, because human has different ground truth goals at different stages of the task plan. Ideal interaction between latent human states $Z^h_t$ and latent robot states $Z^r_t$ would lead to seamless human-robot collaboration, where robot motion always matches human intention $G_t$. To achieve seamless collaboration, $G_t$ must be effectively tracked given observed state history of both the human and the robot $X^{h,r}_{1:t}$, e.g., positions of human skeleton and robot end-effector. First, we study intention tracking when $G_t$ is a random variable. Second, we study hierarchical intention tracking when $G_t$ represents an intention hierarchy with an arbitrary number of layers, and apply to the two-level use case in Figure~\ref{fig:intro}(c).
\subsection{Intention Tracking in Human-Robot Collaboration}\label{sec:architecture-generic}
The human intention variable~$G_t$ at each time step is discrete and the cardinality of its sample space is $m$. Bayesian filtering is recursively applied every $T_p$ time steps to obtain the posteriori over~$G_t$ conditioned on observed states~$X_{1:t}^{h,r}$. For simplicity, we assume the intention outcomes $g_{t+1:t+T_p}$ are consistent. This assumption is reasonable in our implementation because $T_p$ time steps are less than 0.2 second. The prediction step of Bayesian filtering is as follows.
\begin{equation}
P(g_{t+T_p} | x_{1:t}^{h,r}) = \sum_{g_{t}} P(g_{t+T_p} | g_{t}) P(g_{t} | x_{1:t}^{h,r})
\end{equation}
We define the transition model for intention dynamics $P(g_{t+T_p} | g_{t})$ as a time-invariant matrix
\begin{equation}\label{eq:intention-transition}
P(g_{t+T_p}|g_{t}) = \begin{cases}
\alpha, & \text{if $g_{t+T_p} = g_{t}$};\\
(1-\alpha)/(m-1), & \text{if $g_{t+T_p} \neq g_{t}$}.
\end{cases}
\end{equation}
where a large $\alpha$ indicates the human is more likely to keep the current intention. The probability of the human shifting to another intention is equivalent. We preserve non-zero probability for intentions representing tasks already performed by the human, because the human may perform the same task again due to failed robot attempts. The update step of Bayesian filtering is as follows.
\begin{equation}\label{eq:update-step}
P(g_{t+T_p} | x_{1:t+T_p}^{h,r})\! \propto\!P(x_{t+1:t+T_p}^{h,r} | x_{1:t}^{h,r}, g_{t+T_p}) P(g_{t+T_p} | x_{1:t}^{h,r})
\end{equation}
We derive the prediction model $P(x_{t+1:t+T_p}^{h,r} | x_{1:t}^{h,r}, g_{t+T_p})$ by the recursive expression below.
\begin{equation}\label{eq:generic-intention-filtering-prediction-recursive}
    \begin{aligned}
        & P(x_{t+1:t+\tau}^{h,r} | x_{1:t}^{h,r}, g_{t+T_p}) \\ 
        &\! = \!P(x_{t+\tau}^{h} | x_{1:t+\tau-1}^{h,r},  x_{t+\tau}^{r}, g_{t+T_p}) P(x_{t+\tau}^{r} | x_{1:t+\tau-1}^{h,r}, g_{t+T_p}) \\ & \quad \,P(x_{t+1:t+\tau-1}^{h,r} | x_{1:t}^{h,r}, g_{t+T_p}) \\
        &\! = \!P(x_{t+\tau}^{h} | x_{1:t+\tau-1}^{h,r},  x_{t+\tau}^{r}, g_{t+\tau}) P(x_{t+\tau}^{r} | x_{1:t+\tau-1}^{h,r}, g_{t+\tau-1}) \\ & \quad \,P(x_{t+1:t+\tau-1}^{h,r} | x_{1:t}^{h,r}, g_{t+T_p}) \\
        &\! = \!P(x_{t+\tau}^{h} | x_{1:t+\tau-1}^{h,r},  x_{t+\tau}^{r}, g_{t+\tau}) P(x_{t+1:t+\tau-1}^{h,r} | x_{1:t}^{h,r}, g_{t+T_p}) \\
    \end{aligned}
\end{equation}
The second equality in Equation~\ref{eq:generic-intention-filtering-prediction-recursive} is because $g_{t+1:t+T_p}$ are assumed the same. The third equality in Equation~\ref{eq:generic-intention-filtering-prediction-recursive} is because we use the maximum likelihood estimate of the intention as input to the downstream robot control pipeline, which leads to a deterministic mapping from the observed state history to the next observed robot state. We apply the recursive expression and get the prediction model.
\begin{equation}
P(x_{t+1:t+T_p}^{h,r} | x_{1:t}^{h,r}, g_{t+T_p})\!\!=\!\!\prod_{\tau=1}^{T_p} \!\!P(x_{t+\tau}^{h} | x_{1:t+\tau-1}^{h,r}, x_{t+\tau}^r, g_{t+\tau})
\end{equation}
\subsection{Extension to Hierarchical Intentions}\label{sec:architecture-hierarchical}
Consider ${G}_t^l$ and ${G}_t^{l+1}$ of the hierarchical intention structure in Figure \ref{fig:graphical-hierarchical}(b). The intention transition models follow Equation~\ref{eq:intention-transition}. We expect $\alpha^{l+1}$ is larger than $\alpha^{l}$ because a higher-level intention is less frequently changed. We derive the prediction model for ${G}_t^{l+1}$ in terms of the counterpart for ${G}_t^{l}$, where $P(g_{t+T_p}^{l} | x_{1:t}^{h,r}, g_{t+T_p}^{l+1})$ is related to $P(g_{t+T_p}^{l} | x_{1:t}^{h,r})$ and prior knowledge of relations between $g_{t+T_p}^{l}$ and $g_{t+T_p}^{l+1}$. 
\begin{equation}\label{eq:prediction-hierarchical}
    \begin{aligned}
        & P(x_{t+1:t+T_p}^{h,r} | x_{1:t}^{h,r},g_{t+T_p}^{l+1}) \\
        &= \sum_{g_{t+T_p}^{l}} P(x_{t+1:t+T_p}^{h,r} | x_{1:t}^{h,r},g_{t+T_p}^{l}) P(g_{t+T_p}^{l} | x_{1:t}^{h,r}, g_{t+T_p}^{l+1}) \\
    \end{aligned}
\end{equation}
The procedure of Hierarchical Intention Tracking is as follows. First, perform intention tracking at a lower level conditioned on a fixed higher level intention to get $P(g_{t+T_p}^{l} | x_{1:t}^{h,r}, g_{t+T_p}^{l+1})$. Second, derive the prediction model at the higher level $P(x_{t+1:t+T_p}^{h,r} | x_{1:t}^{h,r},g_{t+T_p}^{l+1})$ by Equation~\ref{eq:prediction-hierarchical}. Repeat the first and the second steps until we reach the highest level $L$, where we can perform Bayesian filtering without conditioning on any other intentions to get $P(g^L_{t+T_p} | x_{1:t+T_p}^{h,r})$. We can then get the posteriors from higher levels to lower levels.
 
\begin{equation}\label{eq:high-to-low-general}
    \begin{aligned}
        & P(g^l_{t+T_p} | x_{1:t+T_p}^{h,r}) \\
        & = \sum_{g^{l+1}_{t+T_p}} P(g_{t+T_p}^{l} | x_{1:t}^{h,r}, g_{t+T_p}^{l+1}) P(g^{l+1}_{t+T_p} | x_{1:t+T_p}^{h,r}) \\
    \end{aligned}
\end{equation}

We apply hierarchical intention tracking to our use case as in Figure~\ref{fig:intro}(c), which is a two-layer intention hierarchy composed of a low-level task intention $G_t^1$ and a high-level interactive intention $G_t^2$. The sample space of~$G_t^1$ comprises four task-related goals ($i = 1,2,3,4$) and failure recovery (FR). The sample space of~$G_t^2$ comprises cooperation (CO) and coexistence (CE) interactive intentions. Intention transition at both levels are controlled by~$\alpha^1$ and~$\alpha^2$. In our use case, $P(g_{t+T_p}^{1} | x_{1:t}^{h,r}, g_{t+T_p}^{2})$ is as follows.
\begin{equation}\label{eq:eg-hierarhical-intention-flow}
    \begin{aligned}
        P(FR | x_{1:t}^{h,r}, CO) &= 1 \\
        P(i | x_{1:t}^{h,r}, CO) &= 0 \\
        P(FR | x_{1:t}^{h,r}, CE) &= 0 \\
        P(i | x_{1:t}^{h,r}, CE) &= \frac{P(i | x_{1:t}^{h,r})}{\sum_{j=1}^{4} P(j | x_{1:t}^{h,r})}\\
    \end{aligned}
\end{equation}

In addition, when conditioned on the coexistence interactive intention, we can simplify the prediction model by removing the effect of the robot on human motion.
\begin{equation}\label{eq:high-level-intention-1-prediction-model}
    \begin{aligned}
    & P(x_{t+1:t+T_p}^{h,r} | x_{1:t}^{h,r}, G^2_{t+T_p}=CE) \\
    & = \prod_{\tau=1}^{T_p} P(x_{t+\tau}^{h} | x_{1:t+\tau-1}^{h}, G^2_{t+\tau}=CE) \\
    & = P(x_{t+1:t+T_p}^{h} | x_{1:t}^{h}, G^2_{t+T_p}=CE) \\
    \end{aligned}
\end{equation}
\section{Collaborative Robot Architecture}\label{sec:architecture}

\begin{figure}[t]
    \centering
    \includegraphics[width=\linewidth]{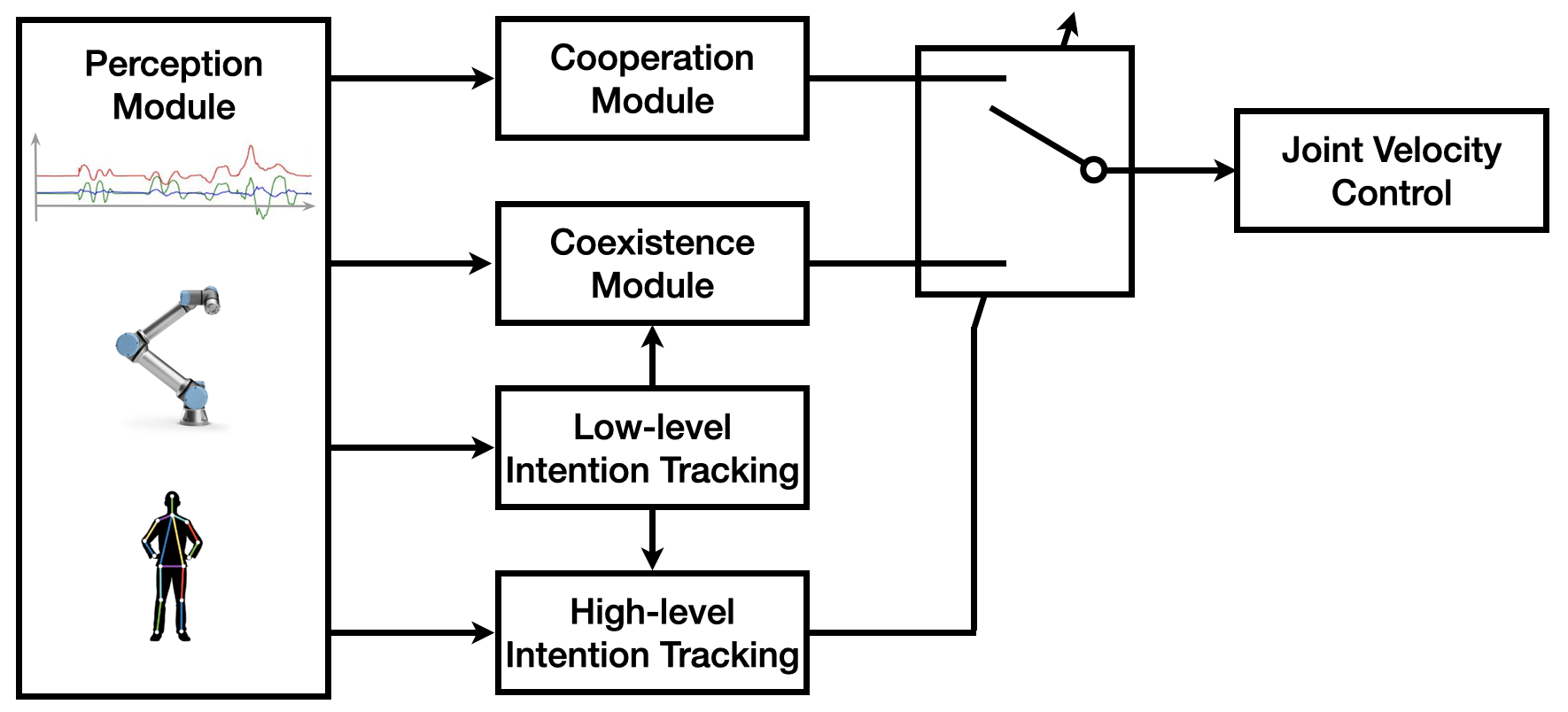}
    \caption{The architecture of Hierarchical Intention Tracking (HIT) based human-robot collaboration system. Perception module includes robot wrist force/torque measurements, robot proprioception, and human skeleton tracking. Low-level intention tracking module takes human wrist positions as input to track task intention. High-level intention tracking module takes human wrist positions, robot end-effector positions, and low-level intention tracking outputs to track interactive intention. Based on the tracked interactive intention, coexistence or cooperation module is selected for motion planning. Coexistence module plans motion according to the task plan generated from the tracked task intentions. Cooperation module plans motion by following human's guidance.}
    \label{fig:pipeline}
    \vspace{-15pt}
\end{figure}

We introduce Hierarchical Intention Tracking (HIT) based human-robot collaboration system as presented in Figure~\ref{fig:pipeline}.

\subsection{Experimental Setup for Collaborative Assembly}\label{sec:architecture-robot}

A human and a robot work together on an assembly task in close proximity. The task involves assembling four pairs of Misumi Waterproof E-Model Crimp Wire Connectors~\cite{kimble2020benchmarking}. The connectors are in asymmetrical shapes and have tight clearances. At the beginning of one experiment trial, female parts are separately placed in four square regions denoting task intentions part $1, 2, 3$, and $4$. All male parts are initially placed in a rectangular region denoting preparation area. The human picks up a male part from the preparation area and aligns it to a female part within the corresponding region. The aligned parts are left at the same region, and the robot reaches to them and perform the pushing action. The human decides the sequence of task intentions for executing part alignment. The robot knows locations of task intention regions, but not the task sequence. The aligned male part may fall off due to table shaking from robot motion. The robot may not push on an ideal position and fail the assembly. The collaborative robot system should robustly handle these cases. 

\subsection{Robot and Perception Setup}\label{sec:architecture-robot}
The robot is a UR5e arm equipped with a Robotiq Hand-E Gripper. An embedded sensor measures force and torque on the end-effector at 500~Hz. The robot is controlled by joint velocity commands and operated at a reduced speed (30\%). Intel RealSense RGBD Cameras provide a top-down and a side view of the shared space to alleviate occlusion. We run OpenPose~\cite{cao2017realtime} to detect human skeleton positions from both views. We apply Kalman Filter to these detections and track 3D positions of human right wrist at 30~Hz. All modules are executed using Robot Operating System (ROS).
\begin{figure}[t!]
    \centering
    \includegraphics[width=0.6\linewidth]{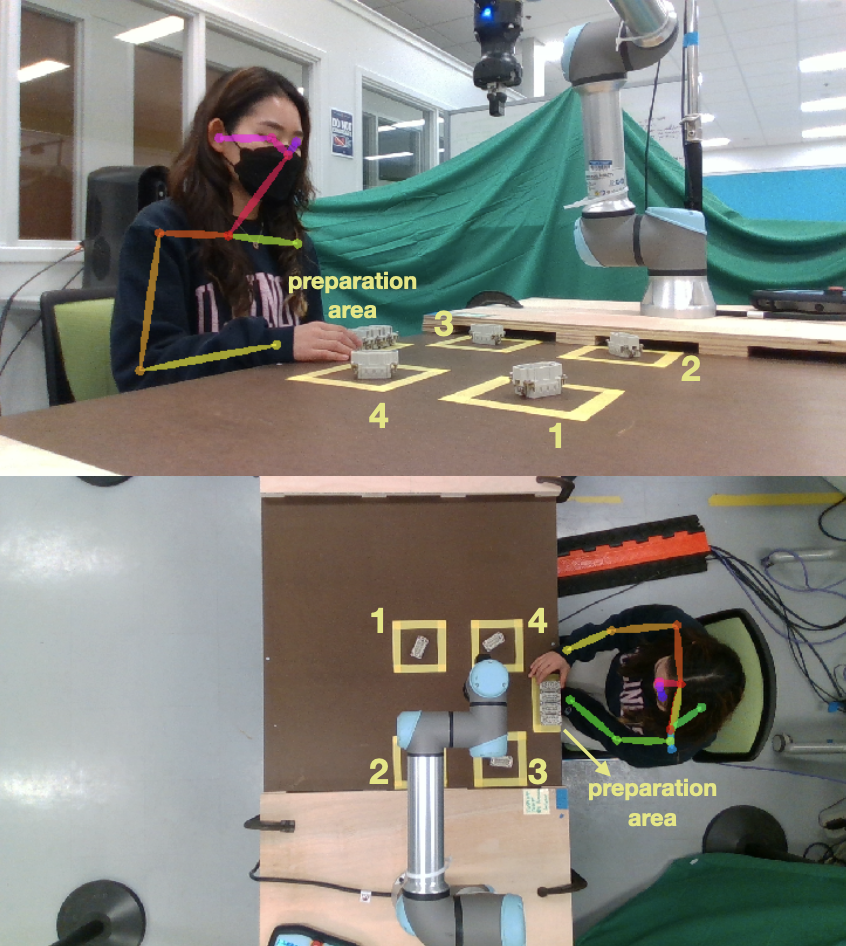}
    \caption{Human skeleton detection by OpenPose on frames from a side-view and a top-view camera. Multiple cameras are used to address occlusion which happens frequently in close-proximity human-robot collaboration.}
    \label{fig:human-tracking}
    \vspace{-15pt}
\end{figure}

\subsection{Low-level Intention Tracking}\label{sec:architecture-low}
Task intentions among four parts and the preparation area are tracked at 30~Hz. We adapt Mutable Intention Filter (MIF) with Intention-aware Linear Model (ILM) as the prediction model~\cite{huang2021long} to the assembly task. MIF is a particle filtering variant for single-layer intention tracking when human is not affected by robot. The inputs are observed 3D human wrist trajectories and potential task intention regions, and the output is the probability distribution over the task intentions.

\subsection{High-level Intention Tracking}\label{sec:architecture-high}
Interactive intentions are tracked at 5~Hz. The low-level intention tracking module feeds $P(g_{t+T_p}^{1} | x_{1:t}^{h,r}, G_{t+T_p}^{2}=CE)$ to Equation~\ref{eq:prediction-hierarchical}, but a probabilistic prediction model for $G_t^1$ is still required to compute the prediction model for $G_t^2$. Thus, we use a Gaussian variant of ILM
\begin{equation}\label{eq:gilm}
    x^h_{t+1} = x^h_{t} + \frac{\Tilde{d}_t}{||g_t^1 - x^h_t||} \left(g_t^1 - x^h_t\right)  + w_t
\end{equation}
by which human wrist moves to the goal region $g_t^1$ at an average speed of $\Tilde{d}_t$ during the observation window, with a Gaussian process noise $w_t$. Note for $g_t^1$ as failure recovery, the goal region is defined as a neighboring region around the robot end-effector position which would move through time.

\begin{table*}[hbt!]
\caption{Results of the ablative pilot study. (*) Completion time of Cooperation Mode Baseline is for reference but does not provide fair comparison against other systems. The robot is implemented at a consistently low speed in Coexistence Mode Baseline or when controlled by the coexistence module of HIT system, in order to enforce safety of human subjects. An effective speed and separation monitoring module~\cite{svarny2019safe} would significantly improve the completion time for Coexistence Mode Baseline and HIT System to match the efficiency of Cooperation Mode Baseline without affecting performance in other metrics. We will leave the development for future work.}\label{table:ablative}
\begin{center}
\setlength\tabcolsep{3pt}
\begin{tabular}{lcccccc}
    \toprule
    System & Completion Time (sec) & Automated Path (m) & Guided Path (m) & Human Force (N) & Human Energy (J) & Number of Failures \\
    \midrule
    Coexistence Mode & 104.2$\pm$17.6 & 3.80$\pm$0.65 & N/A & N/A & N/A & 1.2 \\
    Cooperation Mode & 35.2$\pm$3.4$^{(*)}$ & N/A & 2.36$\pm$0.17 & 5.51$\pm$0.39 & 21.04$\pm$2.52 & 0.0 \\
    HIT System & 97.2$\pm$15.5 & 3.40$\pm$0.51 & 0.65$\pm$0.37 & 0.48$\pm$0.24 & 5.71$\pm$3.30 & 0.0 \\
    \bottomrule
\end{tabular}
\end{center}
\vspace{-10pt}
\end{table*}

\begin{figure*}[t]
    \centering
    \includegraphics[width=0.8\linewidth]{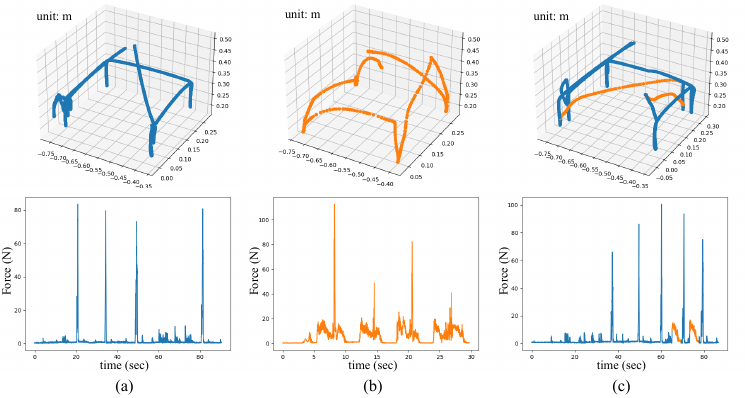}
    \caption{Trajectory and force visualization in (a) Coexistence Mode Baseline; (b) Cooperation Mode Baseline; (c) Hierarchical Intention Tracking system. Blue denotes coexistence mode and orange denotes cooperation mode. Note (c) shows a trial where the robot failed in assembly steps, the human recovered the failure, and the robot performed assembly steps again. The robot did more than 4 assembly steps so there are more than 4 peaks in force visualization in (c).}
    \label{fig:traj-force-plot}
    \vspace{-10pt}
\end{figure*}

\subsection{Planning and Control}\label{sec:architecture-planning}

\textbf{Coexistence Module} is developed for concurrent task execution without interruption. A task set of all parts is initialized. If the most likely task intention belongs to the task set, and its probability stays above 80\% over 1.5 second, the human is detected aligning parts and the intention is moved from the task set to an ongoing task queue. An intention is popped out of the ongoing task queue and pushed into a ready task queue when its probability drops below 25\%, which indicates the alignment in the corresponding region is finished. The robot plans trajectories to reach to and perform the pushing action on the part corresponding to the task intention popped out of the ready task queue. The robot moves repulsively from the human wrist by Artificial Potential Field to avoid collision~\cite{khatib1986real}. Force control is implemented to detect the completion of pushing action. Note a small uniform random noise is added to the goal position of the robot, which is intended to emphasize the robustness by controlling the failure rate of push. 

\textbf{Cooperation Module} is developed for manual guidance to handle failure recovery. If the probability of cooperation interactive intention is above 90\% over 0.5 sec, the robot is switched to the cooperation module. An attractive Artificial Potential Field is implemented so the robot end-effector approaches the human wrist. Admittance control is activated after intended contact is detected, and the human starts guiding the robot to the desired ready-to-push position. Detection on the end of human intervention by force measurement switches the robot back to the coexistence module, and the robot immediately executes the pushing action to recover the assembly failure. A running mean of joint velocities is used to avoid discontinuous control input when switching between coexistence and cooperation modules~\cite{cacace2018shared}.

\section{Ablative Pilot Study}\label{sec:experiments}

\begin{figure*}[t]
    \centering
    \includegraphics[width=\linewidth]{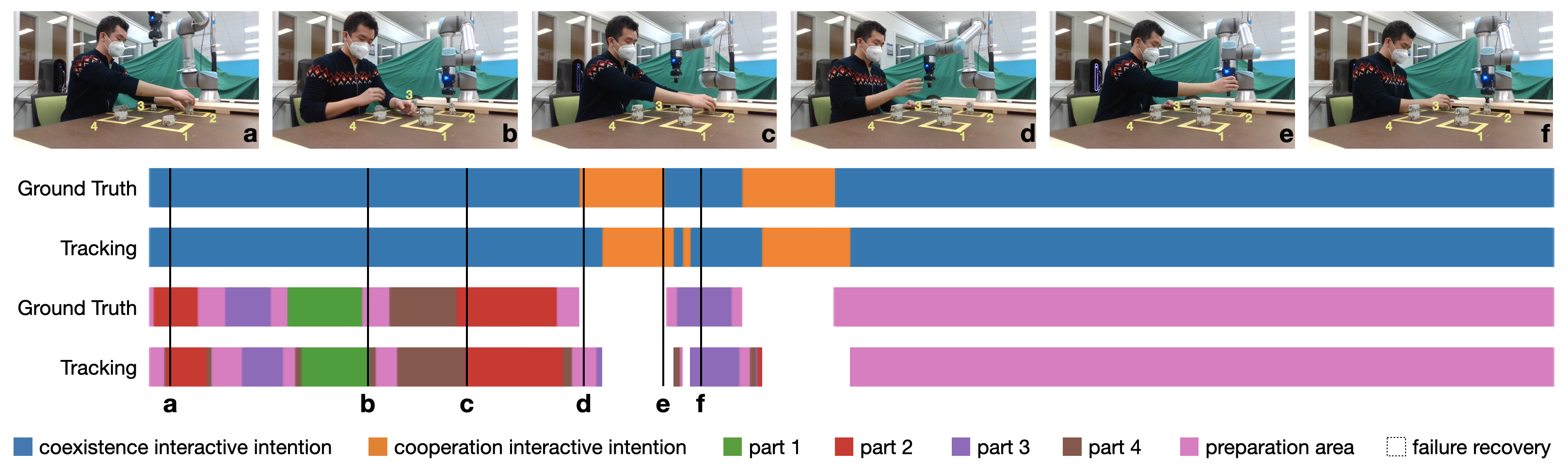}
    \caption{High-level and low-level intention tracking in a trial of the collaborative assembly task. The tracking bar shows the most likely intention tracked at each time step. Snapshots show a failure recovery example. (a) The human aligns parts. (b) The robot fails to assemble the parts. (c) The human realigns the part. (d) The human shows cooperation interactive intention by reaching to the end-effector. (e) The human guides the robot to the appropriate ready-to-push position. (f) The robot recovers the failure by pushing the parts again. The frame-wise accuracy is 90.4\% for low-level intention and 94.5\% for high-level intention.}
    \label{fig:tracking-example}
\end{figure*}

We conduct a pilot study of HIT system against two human-robot collaboration baselines: Coexistence Mode Baseline and Cooperation Mode Baseline. In the Coexistence Mode Baseline, the robot executes low-level intention tracking. The robot is able to perform concurrent task execution with the human but not able to recover the failure. In the Cooperation Mode Baseline, admittance control is executed and the robot is passively compliant. Human guidance is required all the time to move the robot and perform pushing actions. These uni-modal systems are chosen as baselines because coexistence-only or cooperation-only robot are prevalent solutions in industry \cite{villani2018survey, michalos2018seamless}.

Five human subjects participate in the pilot study. Each subject do ten trials of the collaborative assembly task on each system. Evaluation metrics include (1) completion time, (2) length of the automated path executed by the coexistence module, (3) length of the path when robot is guided by the human, (4) average force applied by the human throughout the trial, (5) total energy the human spent throughout the trial, and (6) average number of assembly failures.

The quantitative evaluation is presented in Table \ref{table:ablative}. Coexistence Mode Baseline has limited efficiency in terms of the completion time and the automated path due to two reasons. First, the robot is required to work at a low speed with human in close proximity during the coexistence mode. Second, the robot may not be able to reach an appropriate ready-to-push position for the aligned parts. To prevent assembly failure, the human has to adjust the position of aligned parts multiple times. The robot can get trapped in a local minimum due to human avoidance and goal-directed motion when the human is doing adjustment. There are in average 4.6 times of adjustments in one trial, but there are still 1.2 pairs of parts failed to be assembled in the Coexistence Mode Baseline. Cooperation Mode Baseline has the shortest completion time at the cost of excessive human effort. As demonstrated in Figure \ref{fig:traj-force-plot}(b), the human guides the robot all the time in order to reach aligned parts and perform the pushing actions. Continuous guidance for the goal-reaching motion consumes much human energy and leads to human fatigue, while large impact during the pushing action exhibits poor ergonomics which may cause workplace injuries. 

In contrast to the baselines, HIT system effectively balances coexistence and cooperation to combine advantages from both sides. HIT system has a shorter length of the guided path and significantly decreased human force and energy than the Cooperation Mode Baseline, because human only needs to guide the robot when failure recovery is needed. Figure~\ref{fig:traj-force-plot}(c) demonstrates that the cooperation mode is active only for two short periods, while most goal-reaching motion and all pushing actions are executed by the coexistence module. HIT system also has a shorter length of the automated path compared against Coexistence Mode Baseline, since the human does not have to readjust the part positions with the capability to recover the failure. Though the total length of path is longer than the Coexistence Mode Baseline, HIT system still has a shorter completion time. The robot is allowed to move faster than the coexistence mode when manually guided.

Figure \ref{fig:tracking-example} shows the effective performance of both high-level and low-level intention tracking. In this visualized trial, the human aligns the parts in the order of \{2,3,1,4\}. The assembly of part 2 and part 3 are initially failed and then recovered. The snapshots in Figure~\ref{fig:tracking-example} show the sequence of assembling part 2 including the initial attempt and the failure recovery. Note that high-level and low-level intention tracking modules have not only different tracking frequencies, but also different sensitivities on evolving intentions. The low-level intention tracking has a higher sensitivity and a shorter tracking delay. This short delay is critical to the accuracy of the high-level prediction model which uses the probability distribution over low-level task intentions. The high-level intention tracking has a lower sensitivity so the most likely high-level interactive intention is less likely to jump between coexistence and cooperation, which can result in the instability of robot control.
\section{Conclusions}\label{sec:conclusions}

We propose the concept of hierarchical intention tracking to take into account the continuously changing human intention and its hierarchical structure in the context of human-robot collaboration. We introduce a Hierarchical Intention Tracking (HIT) based human-robot collaboration system which effectively integrates the coexistence and cooperation modules. We demonstrate seamless interaction, robust failure recovery and enhanced ergonomics of the HIT system against baselines through real-world experiments on a collaborative assembly task. In future work, we will develop speed and separation monitoring to improve the efficiency of the coexistence module. We attempt to generalize our framework from rule-based intentions to latent human states learned from variational inference or MCMC, and compare performance with these learning-based methods~\cite{losey2022learning, zolotas2022disentangled}. We plan to perform extensive human subject experiments. We would like to explore a time-varying intention transition setting, such as how to incorporate the prior knowledge of the assembly task sequence by learning from task recordings.

\bibliographystyle{IEEEtran}
\bibliography{bib}

\end{document}